\documentclass{article}

     \usepackage[final]{neurips_2019}

\usepackage{times}  
\usepackage{helvet} 
\usepackage{courier}  
\usepackage[hyphens]{url}  
\usepackage{graphicx} 
\usepackage{graphicx}  
\usepackage[utf8]{inputenc}
\usepackage[T1]{fontenc}
\usepackage{lmodern}
\usepackage{subfigure}
\usepackage{amsmath}
\usepackage{booktabs}
\usepackage{multirow}
\usepackage{newtxtext,newtxmath}
\usepackage[colorlinks=true,citecolor=red,linkcolor=black]{hyperref}

\title{Building Information Modeling and Classification by Visual Learning At A City Scale}

\author{
  Qian Yu \\
  UC Berkeley / ICSI\\
  \texttt{qianyu1023@berkeley.edu} \\
  \And
  Chaofeng Wang$^*$ \\
  University of California, Berkeley\\
  \texttt{c\_w@berkeley.edu} \\
  \And
  Barbaros Cetiner \\
  University of California, Los Angeles\\
  \texttt{bacetiner@ucla.edu} \\
  \And
  Stella X. Yu \\
  UC Berkeley / ICSI\\
  \texttt{stellayu@berkeley.edu} \\
  \And
  Frank Mckenna \\
  University of California, Berkeley\\
  \texttt{fmckenna@berkeley.edu} \\
  \And
  Ertugrul Taciroglu \\
  University of California, Los Angeles\\
  \texttt{etacir@g.ucla.edu} \\
  \And
  Kincho H. Law  \\
  Stanford University\\
  \texttt{law@stanford.edu} \\
}

\begin{document}

\maketitle

\begin{abstract}
In this paper, we provide two case studies to demonstrate how artificial intelligence can empower civil engineering. In the first case, a machine learning-assisted framework, BRAILS, is proposed for city-scale building information modeling. Building information modeling (BIM) is an efficient way of describing buildings, which is essential to architecture, engineering, and construction. Our proposed framework employs deep learning technique to extract visual information of buildings from satellite/street view images. Further, a novel machine learning (ML)-based statistical tool, \textit{SURF}, is proposed to discover the spatial patterns in building metadata. 

The second case focuses on the task of soft-story building classification. Soft-story buildings are a type of buildings prone to collapse during a moderate or severe earthquake. Hence, identifying and retrofitting such buildings is vital in the current earthquake preparedness efforts. For this task, we propose an automated deep learning-based procedure for identifying soft-story buildings from street view images at a regional scale. We also create a large-scale building image database and a semi-automated image labeling approach that effectively annotates new database entries. Through extensive computational experiments, we demonstrate the effectiveness of the proposed method.

\end{abstract}

\section{Introduction}

Natural disasters often bring huge losses to human society. One of the main consequences is the destruction of buildings, usually accompanied by casualties and loss of property. As the major component of a human-built environment, buildings are an important consideration in disaster prevention, response, and reconstruction, on which researchers have made significant efforts. 

In recent years, the computer vision (CV) community has embraced notable improvement. Deep learning (DL), as a part of machine learning (ML), is arguably one of the most popular research topics nowadays. A typical deep convolutional neural network (CNN) is a stack of convolutional layers and fully connected layers. Except for the last layer, each layer serves as a feature extractor. Compared with traditional methods for CV tasks, which require researchers to select the `optimal' feature representation, a CNN model can instead learn to extract features by itself. A series of CNN-based models have been proposed and achieved impressive results in a variety of CV tasks, like face verification \cite{sun2014deep} and object detection \cite{girshick2015fast,ren2015faster}.

In this work, we explore the potential of ML/DL in civil engineering, especially in disaster prevention with regard to buildings. Two case studies are introduced: In the first case, a ML-assisted framework, BRAILS \cite{brails}, is proposed for city-scale building information modeling. Building information model (BIM) is an efficient way of describing buildings. Beyond the conventional modeling methods which rely on metadata such as height and number of stories, we employ a CNN to extract visual features of individual buildings from satellite/street view images to create a more informative BIM database. In addition, a ML-based statistical tool termed \textit{SURF} is brought up to discover the spatial patterns in building metadata, which can further enhance the BIM database. The second case focuses on the application of DL in analyzing seismic damage vulnerability of buildings, particularly soft-story buildings. Such buildings are prone to collapse even in a moderate earthquake due to their structure. We propose a DL-based procedure for identifying soft-story buildings from street view images at a regional scale. Based on our newly collected database, we demonstrate the effectiveness of the proposed method. Next, we will explain each case in details.



\section{Case Studies}
\subsection{Create a City-Scale BIM Database}

\begin{figure*}
\centering
\includegraphics[width=0.95\textwidth]{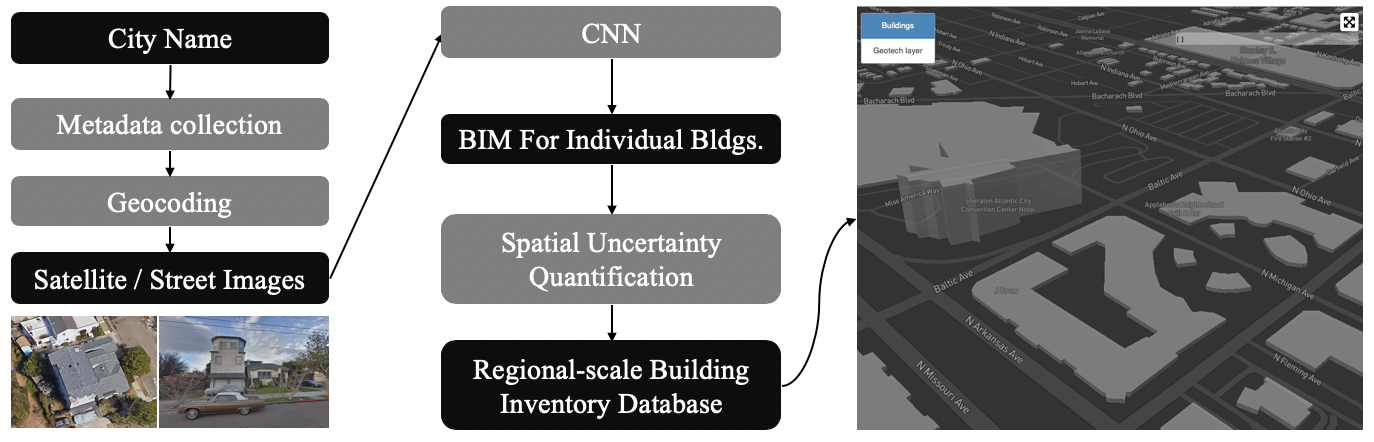}
\caption{Workflow of creating a city-scale BIM database.}
\label{fig:framework}
\end{figure*}

\noindent\textbf{Framework}

The workflow of the current version of BRAILS \cite{brails} for creating a city-scale BIM database is shown in Fig.~\ref{fig:framework}. 

\begin{itemize}
\item \textbf{Step 1:} Given a city of interest, scrape building metadata from property tax website, which contains building information such as building address and number of stories. 
\item \textbf{Step 2:} Associate each scraped building with a geocode (i.e., latitude and longitude) in order to get the precise location of individual buildings. 
\item \textbf{Step 3:} Following the geocodes, fetch images of each building and use CNN to extract visual information from these images. Building information obtained in the first and third step form the skeleton of the BIM database. 
\item \textbf{Step 4:} a ML-based spatial uncertainty quantification tool, SURF, is proposed to enhance the database. It can predict missing values based on the spatial patterns found in the skeleton. At the end of the pipeline, a regional BIM database is created. 
\end{itemize}

\noindent\textbf{Collect Metadata} \quad Property tax records are public resources available on various government websites. Useful information for describing a building can be extracted from these records, such as number of stories, exterior structural material, and year of construction. Based on the extracted information, we can create a preliminary BIM database for each building. This BIM database is called the metadata-BIM database. Note that a considerable portion of the building information is unavailable in the public records. In the final step, a ML-based approach is introduced to compensate for these missing values.

\noindent\textbf{Geocoding} \quad The process of geocoding is to associate individual building (properties) with its geographic coordinates (latitude and longitude). This step is necessary: First, the coordinates are the `index' when retrieving satellite or street view images for individual buildings; Second, in order to enhance the incomplete database (to be explained later), the geographic coordinates will be used for spatial distribution analysis. Google Geocoding API is employed to retrieve the geographic coordinates. 

\noindent\textbf{Extract visual features from building images using CNN} \quad
In addition to the building information obtained from property tax records, a CNN model is utilized to predict building attributes based on their visual features present in satellite or street view images. We train CNNs based on building-related attributes which are collected from OpenStreetMap (OSM)\footnote{OSM is a platform hosting real-world infrastructure information labeled by humans.}. The pipeline of extracting a specific building property from images is listed below:

\begin{itemize}
\item Identify an attribute (e.g., exterior construction material) that is intended to be extracted. 

\item Retrieve satellite/street view images via Google Map API.

\item Label the retrieved images using tags obtained from OpenStreetMap.

\item Train a CNN on labeled images.

\item Use the trained CNN to predict the attribute for unlabeled images.

\end{itemize}

Through repeating the above steps for each building attribute, we create a vision-BIM database. It can be merged with the metadata-BIM to form a more informative BIM database. In later versions of BRAILS, all building features will be obtained using the vision-based method, because scarping metadata from websites is a time consuming task and can not be automated.


\noindent\textbf{Enhance database by exploring spatial correlation} \quad Based on previous steps, a BIM database can be created. However, there is a considerable portion of building information missing in the metadata- and vision-based BIM database. 
Therefore, a ML-based approach termed \textit{SURF} \cite{surf}, which stands for \textit{S}patial \textit{U}ncertainty \textit{R}esearch \textit{F}ramework, is brought up to enhance the database by exploiting spatial patterns in building distribution.

\begin{figure*}
\centering
\includegraphics[width=0.8\textwidth]{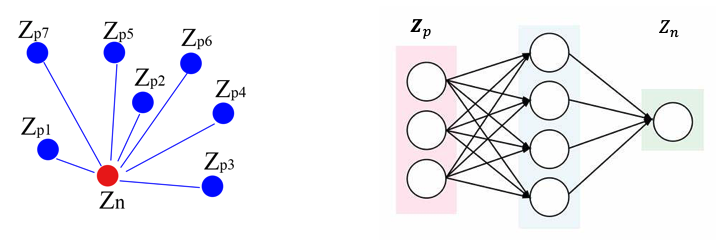}
\caption{Illustration of how SURF learns a mapping function.}
\label{fig:surf}
\end{figure*}

SURF is powered by two engines: random fields \cite{vanmarcke2010random} and neural networks. As shown in Fig. \ref{fig:surf} (Left), the red dot \textit{$Z_{n}$} represents a building with its location known but its property unknown. The blue dots \textit{$Z_{pi}$} represent nearby buildings with their locations and properties known. We can use either random field theory or neural network to construct a mapping function, which is used to map neighbour information \textit{$Z_{pi}$} into \textit{$Z_{n}$}. SURF learns this mapping function and predicts the missing values based on known values of neighbour buildings. Hence, the BIM database is enhanced. More theoretical information of SURF can be found in \cite{surf}, where there are links to its GitHub, documents, etc.

\begin{figure*}
\centering
  \includegraphics[width=0.95\linewidth]{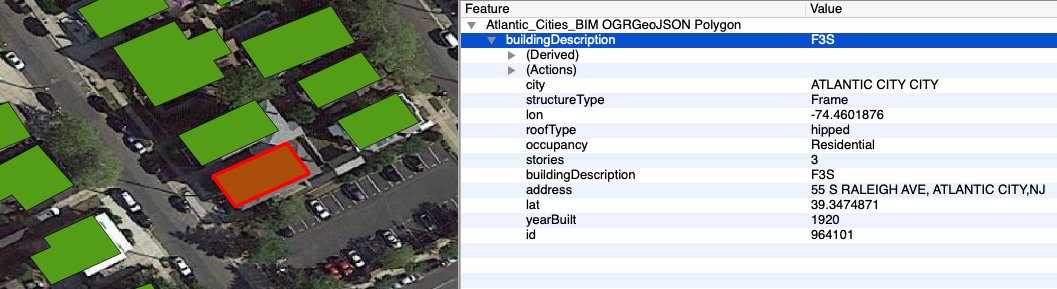}
  \caption{An example of the created BIM database.}\label{fig:BIM_demo}
\end{figure*}

\noindent\textbf{Application example }

Following the framework introduced above, we can create a BIM database based on building information.  Fig.~\ref{fig:BIM_demo} shows an example building with its information in the created BIM database. A BIM database can be used for estimating regional loss in natural hazards, such as earthquakes and hurricanes. An example application has been conducted by \cite{wael_elhaddad_2018_1442914}. 



\subsection{Train a Soft-story Building Classifier}
\label{sec:workflow}
\begin{figure*}
\centering
\includegraphics[width=\textwidth]{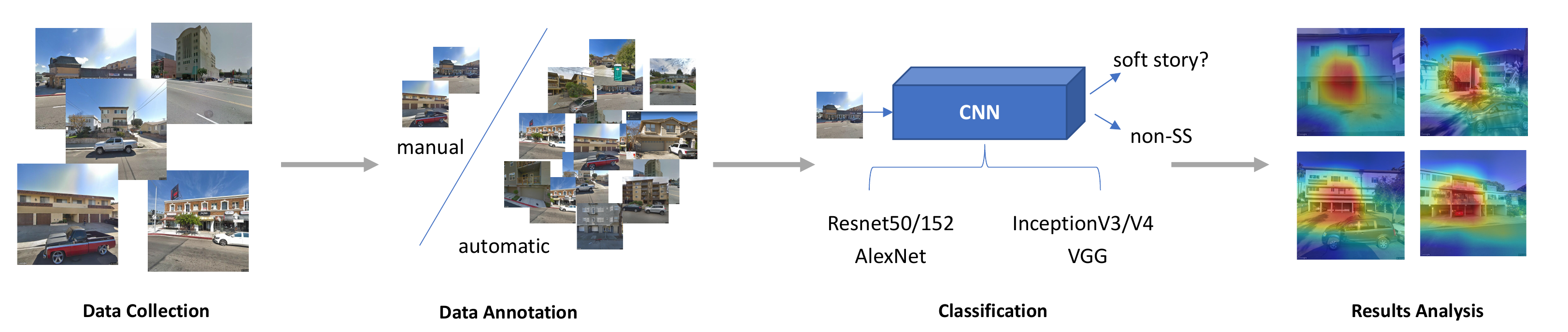}
\caption{Overall workflow of the proposed framework.}
\label{fig:pipeline}
\end{figure*}

In the second case, we propose a deep learning-based framework for soft-story (SS) building classification. To be specific, we collect a large-scale building database using Google Street View Static API\footnote{Google Street View Static API: https://developers.google.com/maps/documentation/streetview/intro.}. Based on the new database, a CNN model is developed to classify a SS building from a street view image. The database contains 25K images, corresponding to buildings of five cities: Santa Monica, Oakland, San Francisco, San Jose, and Berkeley. Training a classifier requires labels, while annotating such a large amount of images is labor-intensive and time-consuming. To handle this problem, we introduce a semi-automatic labeling strategy. Then we train four advanced deep neural networks on these images. Next we will detail the process of creating the new database and training a classifier, as shown in Fig.~\ref{fig:pipeline}. 

\noindent{\bf Create a Building Database}

The process of collecting a database contains two steps, image collection and annotation. 

\noindent{\bf Image Collection} \quad We first obtain building addresses of each city from their official websites. Based on these addresses, we use Google Street View Static API to download the corresponding street view images of  individual buildings. Parameters of the API, such as field of view, pitch, and heading, are manually set and kept consistent across different cities. Images are grouped into four subsets according to their source cities, with the exception that we merged images of San Jose and Berkeley due to their limited data. 


\noindent{\bf Image Annotation} \quad Manually annotating the whole database can be expensive and time-consuming. Inspired by the idea of active learning, we propose a semi-automatic labeling strategy. The basic idea is training a classifier on a small \textit{manually} annotated subset first, then applying this classifier to categorize a larger subset. This process can be repeated based on requirements. Specifically, we first ask an expert to annotate a small subset of images. Only the images contain \textit{sufficient} and \textit{clear} visual cues are labeled as SS buildings. Based on annotated data, a balanced dataset $D_{m}$ is formed. Next, a CNN model is trained based on $D_{m}$. Once the classifier $f_{p}(.)$ is trained, it is then used to classify the rest of unlabeled data. In our case, we randomly sample 1,302 images from Santa Monica as $D_{m}$. An ImageNet pre-trained ResNet50 is employed as $f_{p}(.)$. 

The preliminary classifier $f_{p}(.)$ can achieve 86.9\% accuracy on $D_{m}$. Next, we apply it to categorize the rest images of Santa Monica (15K) and Oakland (1.3K). We admit that the annotated data still contain noise while the quality can be further improved by conducting this labeling process repeatedly. The final version of the new database contains 25,340 building images. 


\noindent{\bf Train a Building Classifier}

Our goal is to develop a classifier which can recognize a SS building in a street view image. There are two classes in our database, SS and non-SS building; thus, we employ a cross-entropy loss to train our model, as shown in Eq.~\ref{equ:binary}. ${p}_k(x)$ and $l(x)$ are the predicted probability of an input image $x$ and its ground truth label. We take advantage of the existing large-scale auxiliary database by using an \textit{ImageNet} \cite{imagenet_cvpr09} pre-trained model for initialization. The number of output digits is changed from 1000 to 2 to fit our task. During training, we first fine-tune the new fully connected layer while all previous layers are frozen, and then fine-tune all layers. It can help to speed up the convergence. We select four popular CNNs with various architectures or depths as the backbones and compare their performance in Table \ref{tab:SantaMonica}.

\begin{equation}
\label{equ:binary}
L(x) = \sum_{k=1}^{2} {-l(x)\log({p}_k(x))},
\end{equation}

\begin{figure*}[!htb]
\minipage{0.6\textwidth}
  \includegraphics[width=\linewidth]{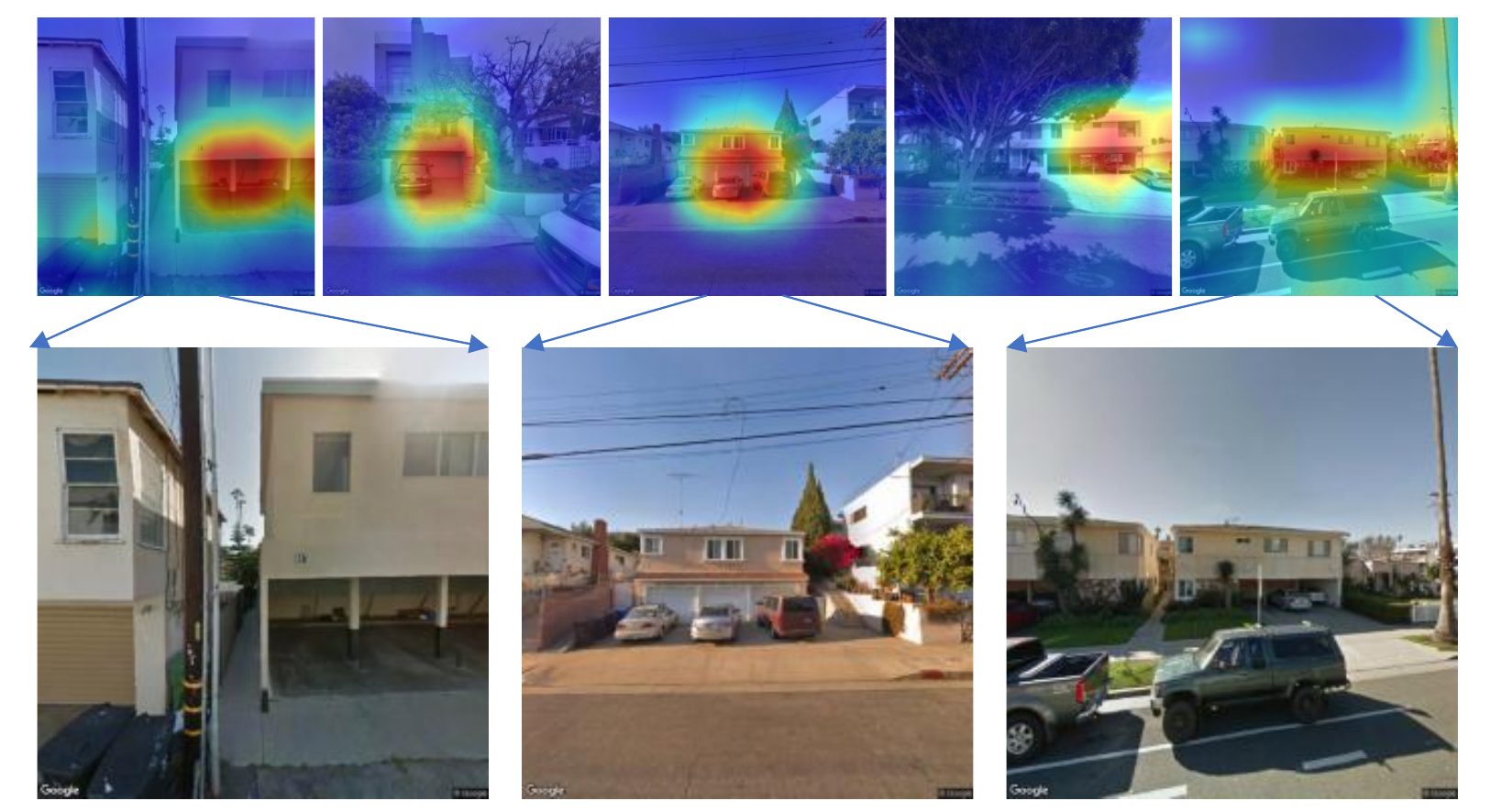}

\endminipage\hfill
\minipage{0.4\textwidth}
  \includegraphics[width=\linewidth]{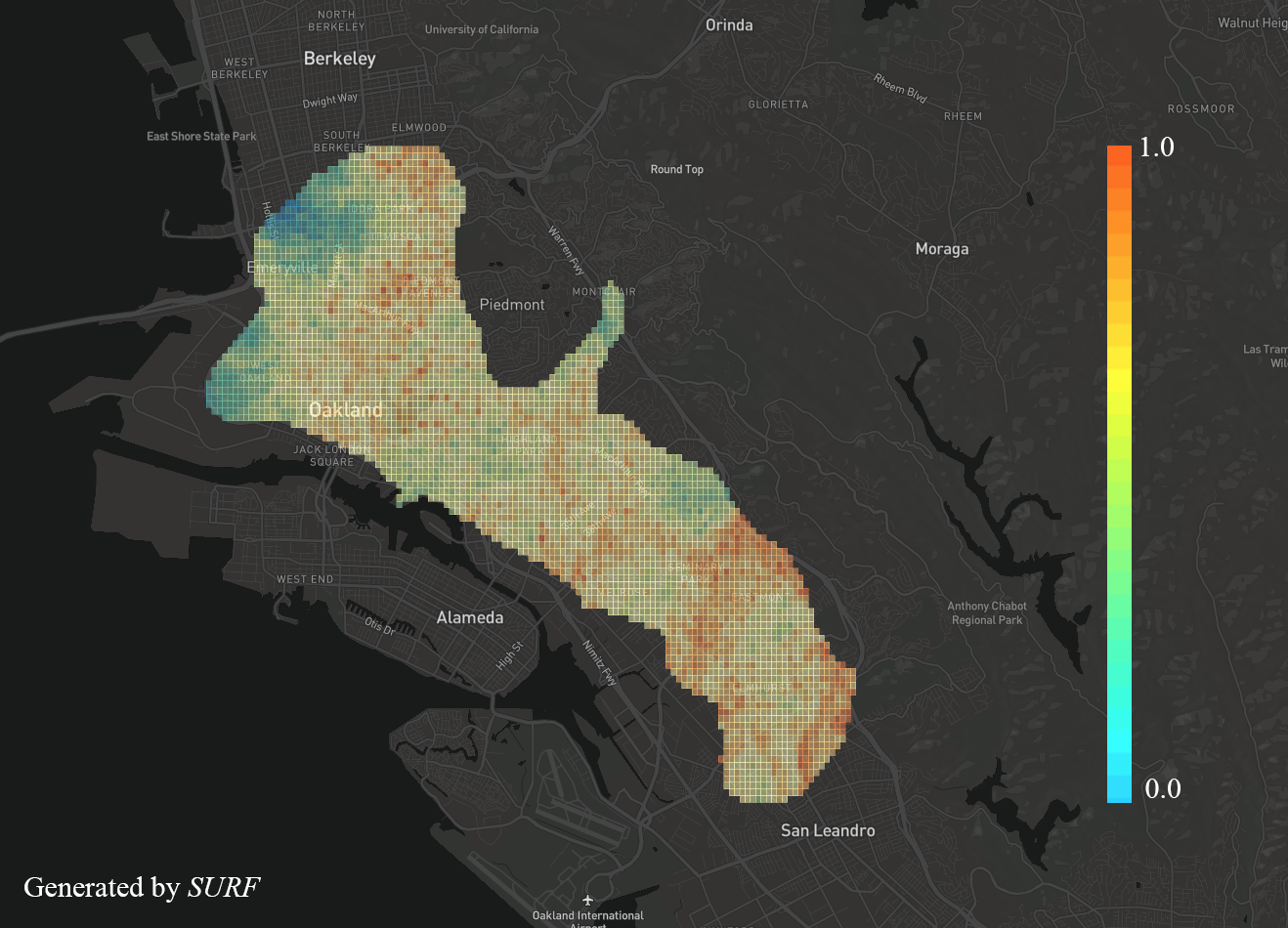}

\endminipage\hfill
\caption{Left: Visualizations of CAM (top) on correctly classified images (bottom) ; Right:Predicted soft-story building distribution in Oakland. The value means the probability of being identified as a soft-story building.}\label{fig:heatmap}
\end{figure*}

\noindent\textbf{Visualizations} \quad
We employ CAM \cite{zhou2016learning} to visualize feature maps learned by our model. CAM is a visualization method; given a target class, CAM highlights the class-specific discriminative regions. We can use this method to see the regions used by the trained model to make a prediction. Figure \ref{fig:heatmap} (Left) provides several examples. In the top row of Fig.~\ref{fig:heatmap} (Left), the highlighted parts indicate regions the model uses to predict a soft-story building. It is clear to see our model attends to building area when predicting whether it is a soft-story or not. This observation is in alignment with \cite{zhou2016learning}, and our trained model can be used as a building detector. 

\begin{table}[tbp]
\centering
\caption{Performance of four networks on Santa Monica subset.}
\begin{tabular}{c|c|ccc}
\toprule[1.4pt]
Model & average acc. & P & R & F1 \\            
\hline
ResNet50\cite{he2015resnet} & \textbf{85.94}\% & \textbf{84.16}\% & 82.80\% & \textbf{83.47}\%   \\
ResNet152\cite{he2015resnet} & 85.03\% & 82.32\% & 83.12\% & 82.71\%  \\
InceptionV3\cite{szegedy2016inception} & 84.38\% & 81.39\% & \textbf{83.77}\% & 82.56\%   \\
InceptionV4\cite{szegedy2017inception} & 83.20\% &  80.52\% & 80.52\% & 80.52\%   \\
\bottomrule[1.4pt]
\end{tabular}
\label{tab:SantaMonica}
\end{table}

\noindent\textbf{Application} \quad The goal of the proposed framework is to provide a prediction given an input street view image. When city-wide images are available, this framework can efficiently process large amounts of data, and the predictions can be used to generate a Soft-story distribution map. 

Here we take the city Oakland as an example. In our collected database, there are 1,359 street view images taken in Oakland. Based on these images and their predictions, we generate a distribution map of SS buildings using \textit{SURF}, which is introduced in the first case study. As shown in Fig. \ref{fig:heatmap}, a heat map is produced to show which parts of the region are likely to be occupied by SS buildings.

\section{Conclusion}

This work demonstrates how ML/DL can benefit civil engineering by introducing two case studies, one being a ML/DL-based approach of creating a BIM database, and the other being a DL-based procedure of creating a database and training a classifier. Apart from this work, other works such as \cite{naik2014streetscore,law2018take,kang2018building,gebru2017using} have applied DL in various applications, like predicting housing price \cite{bency2017beyond,law2018take}, evaluating the safety of the neighborhood\cite{naik2014streetscore,liu2017place}, and estimating the demographic makeup of neighborhoods \cite{gebru2017using}. The impressive results achieved by DL in these applications demonstrate its great potential in civil engineering.

Based on this work, for the first case study, we will explore how to use DL/ML techniques to predict geometric parameters based on a single image. For the second case, we will focus on several practical problems, such as data imbalance and data noise. Moreover, we will also investigate the task of multi-class building classification.

OSM is more complete in higher income countries, densely urbanized areas, and targeted areas of humanitarian mapping intervention, comapred to low and middle income contries and regions. Google maps (street and satellite images) have similar issue. Since the current version of the framework utilizes data from OSM and Google maps, the genealizability of this framework requires more validation work to be performed outside the data rich zones.

Both cases are parts of a larger effort, where the objective is to detect the features of buildings from images at a large scale. To this end, the BRAILS \cite{brails} project is now a hub for them and the development is still on going.



\bibliographystyle{unsrt}

\end{document}